\documentclass[letterpaper]{article} 
\usepackage{aaai25}  
\usepackage{times}  
\usepackage{helvet}  
\usepackage{courier}  
\usepackage[hyphens]{url}  
\usepackage{graphicx} 
\usepackage{natbib}  
\usepackage{caption} 
\usepackage{adjustbox}
\usepackage{multicol}
\usepackage{multirow}
\usepackage{booktabs} 
\usepackage{xcolor}
\usepackage{colortbl}
\usepackage{enumitem}
\usepackage{overpic}
\usepackage{subcaption}
\usepackage{amsmath}
\usepackage{amssymb}
\usepackage{mathtools}
\usepackage{amsthm}

\usepackage{bm}
\usepackage{pifont}
\usepackage{amsfonts}
\usepackage{diagbox}
\usepackage[ruled]{algorithm2e}
\usepackage{algorithmic}

\makeatletter
\DeclareRobustCommand\onedot{\futurelet\@let@token\@onedot}
\def\@onedot{\ifx\@let@token.\else.\null\fi\xspace}

\def\eg{\emph{e.g}\onedot} 
\def\ie{\emph{i.e}\onedot}

\makeatother

\newcommand{\Amat}{\bm{A}}
\newcommand{\Bmat}{\bm{B}}
\newcommand{\Cmat}{\bm{C}}

\newcommand{\dmark}{\ding{51}}

\newcommand{\R}{\mathbb{R}}

\newcommand{\E}{\mathbb{E}}

\newcommand{\mG}{\mathcal{G}}
\newcommand{\mV}{\mathcal{V}}
\newcommand{\mS}{\mathcal{S}}
\newcommand{\mE}{\mathcal{E}}
\newcommand{\mD}{\mathcal{D}}
\newcommand{\mF}{\mathcal{F}}
\newcommand{\mL}{\mathcal{L}}
\newcommand{\mP}{\mathcal{P}}

\newcommand{\mA}{\mathcal{A}}
\newcommand{\mC}{\mathcal{C}}
\newcommand{\mU}{\mathcal{U}}
\newcommand{\mM}{\mathcal{M}}

\usepackage{newfloat}
\usepackage{listings}
\DeclareCaptionStyle{ruled}{labelfont=normalfont,labelsep=colon,strut=off} 
\title{MOL-Mamba: Enhancing Molecular Representation with  \\  Structural \& Electronic Insights
}

\author{
    Jingjing Hu\textsuperscript{\rm 1}\equalcontrib,
    Dan Guo\textsuperscript{\rm 1,2}\equalcontrib\thanks{Corresponding authors},
    Zhan Si\textsuperscript{\rm 3}\equalcontrib,
    Deguang Liu\textsuperscript{\rm 4}, 
    Yunfeng Diao\textsuperscript{\rm 1},
    Jing Zhang\textsuperscript{\rm 1},  \\
    Jinxing Zhou\textsuperscript{\rm 1}, 
    Meng Wang\textsuperscript{\rm 1†}
}
\affiliations{
    \textsuperscript{\rm 1}School of Computer Science and Information Engineering, Hefei University of Technology\\
    \textsuperscript{\rm 2}Institute of Artificial Intelligence, Hefei Comprehensive National Science Center\\
    \textsuperscript{\rm 3}Department of Chemistry and Centre for Atomic Engineering of Advanced Materials, Anhui University\\
    \textsuperscript{\rm 4}Department of Applied Chemistry, 
    University of Science and Technology of China\\
    xianhjj@mail.hfut.edu.cn, guodandd@gmail.com, naa0528@163.com, ldg123@mail.ustc.edu.cn, \\
    \{sjdiaoyunfeng, hfutzhangjing, zhoujxhfut, eric.mengwang\}@gmail.com
}

\usepackage{bibentry}

\begin{document}

\maketitle

\begin{abstract}
Molecular representation learning plays a crucial role in various downstream tasks, such as molecular property prediction and drug design. To accurately represent molecules, Graph Neural Networks (GNNs) and Graph Transformers (GTs) have shown potential in the realm of self-supervised pretraining. However, existing approaches often overlook the relationship between molecular structure and electronic information, as well as the internal semantic reasoning within molecules. This omission of fundamental chemical knowledge in graph semantics leads to incomplete molecular representations, missing the integration of structural and electronic data. To address these issues, we introduce MOL-Mamba, a framework that enhances molecular representation by combining structural and electronic insights. MOL-Mamba consists of an Atom \& Fragment Mamba-Graph (MG) for hierarchical structural reasoning and a Mamba-Transformer (MT) fuser for integrating molecular structure and electronic correlation learning. Additionally, we propose a Structural Distribution Collaborative Training and E-semantic Fusion Training framework to further enhance molecular representation learning. Extensive experiments demonstrate that MOL-Mamba outperforms state-of-the-art baselines across eleven chemical-biological molecular datasets. 

\end{abstract}

\begin{links}
\link{Code}{https://github.com/xian-sh/MOL-Mamba}
\end{links}

\section{Introduction}
\label{sec:intro}
Molecular property prediction is a critical task in various fields, including drug discovery, material science, and chemical engineering. Accurate molecular representation is essential for predicting molecular properties and facilitating the design of novel molecules. Self-supervised pretraining methods~\cite{aaai_geomgcl, aaai_kcl, nips_hp, iclr_graphmvp} have shown significant promise in this area, as they leverage vast amounts of unlabeled data to learn useful representations without the need for extensive human annotations. These methods can capture complex patterns and relationships within molecular structures, making them highly valuable for downstream tasks.

Early molecular representation is feature-driven, with research focusing on the computation of chemical molecular fingerprints and descriptors~\cite{jcim_ecfp, data_moleculenet}, where molecular descriptors are numerical molecular chemical constants that can be quickly obtained by mathematical logic calculations~\cite{moriwaki2018mordred}. These descriptors include quantitative physical, chemical, or topological features of the molecule and are essential for summarizing our understanding of molecular behavior. 
In recent years, data-driven molecular representations have received great popularity due to the rapid development of graph neural networks (GNNs)~\cite{iclr_pretraingnn} and Transformer sequence models (Transformers)~\cite{nips_transformer}. 
Largely, the data-driven molecular pretraining focus on learning representations from molecular structures. Methods such as PretrainGNN~\cite{iclr_pretraingnn}, GROVER~\cite{nips_grover}, and GEM~\cite{nmi_gem} have utilized GNNs and self-supervised learning techniques to explore 2D topologies and 3D spatial information of molecules. These approaches have achieved notable success by predicting node attributes, graph motifs, and geometric features, thereby enhancing the structural representation of molecules. 
However, this does not mean that researchers discard the study of features such as molecular descriptors; on the contrary, molecular representation has evolved towards a more multimodal fusion. 
KCL~\cite{aaai_kcl} and TDCL~\cite{nips_tdcl} facilitate the fusion of expression with molecular structure information by introducing chemical feature knowledge, thus significantly enhancing the molecular learning of the model. 

Currently, the methodological study of structural and electronic feature of fusion molecules as an emerging trend still has a lot of room for development, and this paper adapts to this trend and further pushes it forward. 
Technically, most of the existing molecular representation methods are based on GNNs, and Graph Transformer architectures (GTs)~\cite{nips_graphformer}. 
While GNNs excel in capturing local structures, they often struggle with long-range dependencies due to over-squashing~\cite{hu2024unified}. 
GTs can effectively model global relationships but lack the necessary inductive biases for graph structures.
Inspired by the power of state-space models~\cite{arxiv_stg-mamba}, such as mamba~\cite{gu2023mamba}, a powerful successor to Transformers, for sequential causal inference. In our work, we combine the strengths of both GNNs and GTs by introducing Mamba-enhanced graph learning. This novel approach leverages Mamba's ability to retain contextual information across distant nodes, addressing the limitations of conventional GNNs. By integrating Mamba, we design a graph learning framework that significantly improves the model's capacity to capture complex molecular structures. 
During the fusion of structural and electronic features, we employ a hybrid mechanism that combines Mamba with traditional attention techniques. This integration allows for a more comprehensive representation of molecular properties, effectively balancing the strengths of local and global information processing. By incorporating Mamba's advanced capabilities, our architecture enhances the model's performance in accurately predicting molecular properties, paving the way for more effective applications in drug discovery and material science.

Specifically, we introduce MOL-Mamba (Molecular Mamba), a novel framework designed to enhance molecular representation by integrating structural and electronic insights. MOL-Mamba consists of an Atom \& Fragment Mamba-Graph (MG) for hierarchical structural reasoning and a Mamba-Fusion Encoder (MF) for combining molecular structure and electronic correlation learning. Our approach also incorporates a Structural Distribution Collaborative Training and E-semantic Fusion Training framework to further refine molecular representation learning. By implementing internal semantic reasoning at the fragment and atom levels and fusing structural and electronic data, MOL-Mamba provides a more comprehensive understanding of molecular properties. Our contributions are threefold: 

\begin{itemize}[leftmargin=*]
    \item We propose MOL-Mamba, a new Mamba-enhanced framework that integrates structural and electronic information for improved molecular representation learning.

    \item We introduce novel training strategies, including Structural Distribution Collaborative Training and E-semantic Fusion Training, to enhance the fusion of molecular data.

    \item Extensive experiments demonstrate that MOL-Mamba outperforms state-of-the-art baselines on multiple molecular datasets, providing superior performance and interpretability in molecular property prediction tasks.
\end{itemize}

\section{Related Work}

\subsection{Molecular Representation Learning}
Molecular representation learning, as the foundation for molecular property prediction, has always been a popular research area. Early feature-driven methods included studies on molecular fingerprints~\cite{jcim_ecfp, cs_ecfp}, electronic \& structural descriptors~\cite{stanton1990development, moriwaki2018mordred}. In recent years, data-driven molecular representation methods have gained more attention, directly leveraging molecular SMILES~\cite{iclr_unimol, bb_smiles}, 2D topological structures~\cite{iclr_gin, iclr_graphmvp}, and 3D geometric conformations~\cite{nips_schnet, iclr_dimenet, aaai_lagnet}. It is worth noting that the current trend does not completely abandon feature-driven approaches; rather, it focuses on constructing more integrated models that fuse multiple molecular modalities~\cite{iclr_unimol}. KCL~\cite{aaai_kcl} introduces the chemical element knowledge into the molecular structural graph for knowledge-enhanced molecule learning. TDCL~\cite{nips_hp} calculates the persistent homology fingerprint (a topological feature of the molecule) as an efficient supervision for pretraining molecular structures. 
In order to facilitate the study of multimodal molecular representation, we work on the mining of molecular multi-view structures and their integration with molecular electronic descriptors.

\subsection{Transformers \& Mambas for Molecule Learning}
Transformers~\cite{nips_transformer, iclr_transformer-m, mm_mask} have achieved great success in sequence modeling. Graph Transformers~\cite{iclr_unimol, acs_data} 
allows each node to attend to all others, enabling to effectively capture long-range dependencies, avoiding over-aggregation in local neighborhoods like MPNNs~\cite{icml_mpnn, nips_dmpnn}. 
Recently, as a powerful successor to Transformers, Mamba~\cite{gu2023mamba} has become more friendly to handling long sequences with a linear scale, and its state space model (SSM) has further enhanced sequence inference capabilities, making it a promising sequence modeling solution. Some works~\cite{arxiv_graphmamba, arxiv_stg-mamba} have combined Mamba with graphs to improve the model's nonlinear modeling ability, particularly in graph-based prediction tasks such as social networks. Motivated by these works, we attempt to introduce a new Mamba-based model namely MOL-Mamba, into molecular graph and sequence tasks, to explore a novel molecular representation method.

\begin{figure*}[t]
\centering
\begin{overpic}[width=1\linewidth]{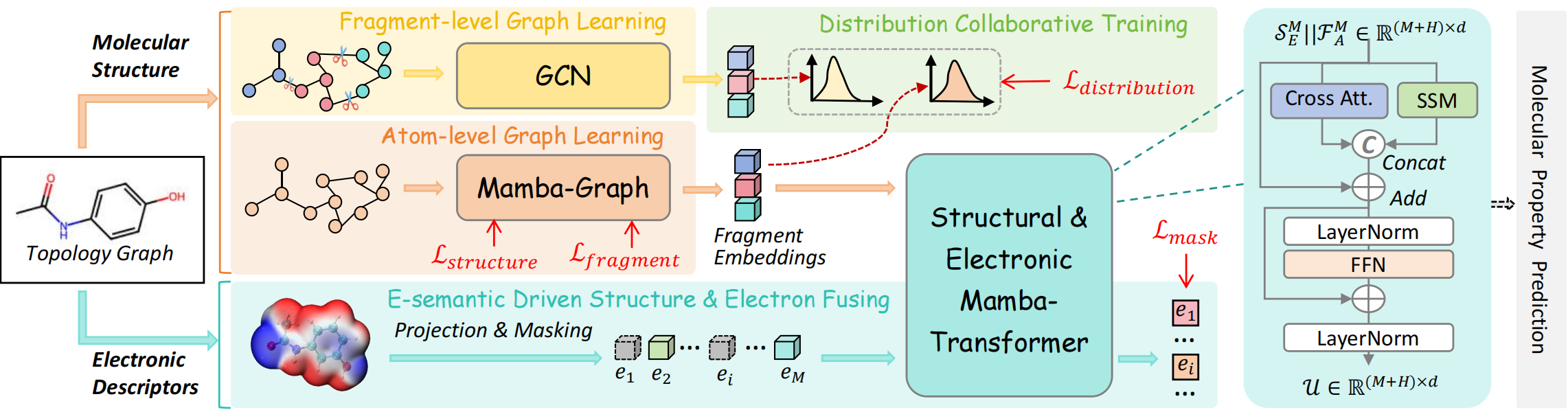}
\end{overpic}
\caption{{\bf The pretraining framework of our approach MOL-Mamba (\underline{Mol}ecular \underline{Mamba}).} It consists of three modules: fragment-level graph \textbf{GNN$_F$}, atom-level structural reasoning Mamba-graph (MG), and a molecular structural \& electronic Mamba-Transformer (MT) fuser. We implement two pretraining stages: molecular structure learning with the Distribution Collaborative Training, then E-semantic Fusion Training is conducted for molecular structural \& electronic fusion learning. 
}
\label{fig:model}
\end{figure*}


\section{Methodology}

\subsection{Problem Formulation}
Generally, a molecule contains both structural and electronic attributes. The molecular structure can be represented as a \emph{atom-level} graph $\mG_A=\{\mV_A, \mE_A, \mC_A\}$, where $\mV_A\in\R^{l\times d^a}$, $\mE_A\in\R^{2\times q}$ and $\mC_A\in\R^{l\times 3}$ denote the node (atom) set, edge (bond) set and the position matrix for atoms, respectively.  
We can mask some edges of the graph $\mG_A$ to generate the molecular fragment subgraphs $\mF=\{\mS_1,\,\mS_2,\cdots\}$. Let these fragments as a node unit, and the inter-fragment bonds as edges, the \emph{fragment-level} graph can be represented as $\mG_F=\{\mV_F, \mE_F\}$. Thus, we have two views of the molecular graphs: {atom-level} graph $\mG_A$ and {fragment-level} graph $\mG_F$. 
The molecule in electron view can be represented as $\mD_E\in\R^M$ with $M$ different electronic descriptors (e-descriptors). 
In the setting of self-supervised molecular representation learning, our goal is to learn graph and e-descriptor encoders $f: \{\mG_A, \mG_F, \mD_E\} \mapsto \R^d$ which maps the input graphs and e-descriptors to a vector representation without any label information. The learned encoders can then be used for various downstream property prediction tasks through finetuning.

\subsection{Bi-Level Graph Collaboration Learning}

\subsubsection*{Fragment-level Graph Construction.} %
Graph fragmentation plays a fundamental role in the quality of the learning models because it dictates the global connectivity patterns. Here, we borrow the Principal Subgraph Mining (PSM) algorithm~\cite{nips_psma} to quickly and efficiently decouple the structure of molecules, where all subgraphs have their unique identifier ids. 
Subsequently, $h$ generated molecular fragment subgraphs $\mF=\{\mS_1,\,\mS_2,\cdots,\mS_h\}$ serve as the nodes $\mV_F$ of fragment-graph $\mG_F$, where $\mS_i=\{\tilde{\mV}^{(i)},\tilde{\mE}^{(i)}\}$ is a fragment subgraph with $\tilde{\mV}^{(i)} \cap \tilde{\mV}^{(j)} = \emptyset,\, \cup^h_{i=1}\tilde{\mV}^{(i)}={\mV}_A$. 
An edge exists between two fragment nodes if there exists at least a bond interconnecting atoms from the fragments. Formally, $\mE_F = \{(i,j) \mid \exists\, u, v, u \in \tilde{V}^{(i)}, v \in \tilde{V}^{(j)}, (u,v) \in \mE_A\}$. After that, the fragment-graph $\mG_F=\{\mV_F, \mE_F\}$ is constructed to explicitly represents the higher-order connectivity between the large components within a molecule, as shown in Figure~\ref{fig:model} (fragment-level graph learning), supplementing molecular structure learning. The embedding of the subgraph's id are used to initialize the fragment node, exploiting the local learning advantage of GNNs~\cite{iclr_gcn, iclr_gin}, we explicitly extract high-order semantics on the structure of molecular fragments.
\begin{align}
    \mF_F = \text{GNN}_F(\text{LN}(\mV_F, \mE_F)) \in \R^{h\times d^f}\,,
\end{align}
where $\text{LN}$ is LayerNorm operation~\cite{nips_transformer}, $d^f$ is the fragment feature dimension.

\subsubsection*{Atom-level Mamba-Graph Construction.} 
\begin{figure*}[t]
\centering
\begin{overpic}[width=0.95\linewidth]{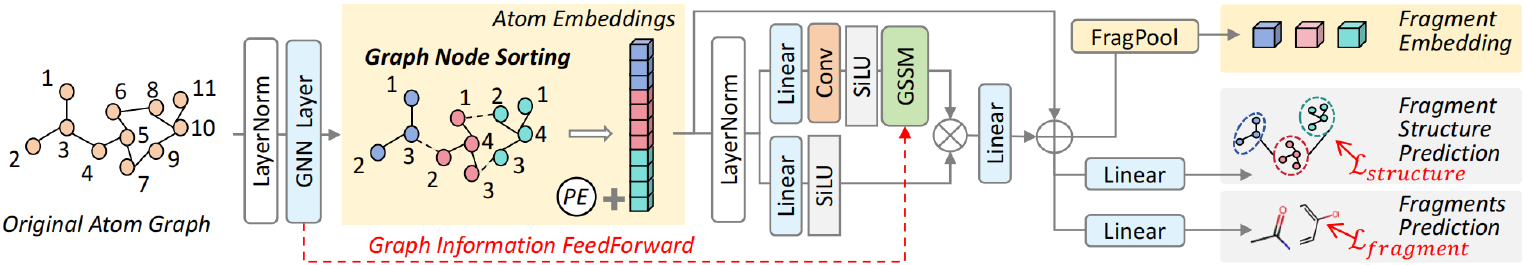} 
\end{overpic}
\caption{{\bf Illustration of the Mamba-Graph (MG) module.} 
It consists of the GNN layer and Mamba block. 
A novel \textbf{graph node sorting strategy} to exploit Mamba's context-aware reasoning. A new \textbf{GSSM mechanism} is designed to adaptively select relevant information from the graph context. Two fragment-related loss constraints $\mL_{structure}$ and $\mL_{fragment}$ facilitate molecular structure learning of the MG module. 
}
\label{fig:mamba-graph}
\end{figure*}

Conventional GNNs directly on the atom-graph suffers from over-squashing and poor capturing of long-range dependencies issues~\cite{arxiv_gm, icml_gred}, especially for molecules with a high atomic number. 
To augment the model's ability to inductively correlate structural contexts within the molecular graph, and inspired by the success of Mamba~\cite{gu2023mamba} in powerful and flexible sequence context modeling, we combine GNN and Mamba namely Mamba-Graph (MG) to learn structural associations within the whole molecule graph, which is more complex compared to explicit fragment structure learning, as shown in Figure~\ref{fig:mamba-graph}. 
Specifically, we use the LN and GNN layer~\cite{nips_mxm} to initialize the node embedding of atom-graph $\mG_A$ with its position information $\mC_A$, then we propose a novel \textbf{graph node sorting strategy}, by considering molecular fragment structure and the node degree information (reflecting the importance of the node), so that the data structure is adapted to exploit Mamba's context-aware reasoning: 
\begin{align}
    &\mF_A^{G} = \text{GNN}_A(\text{LN}(\mV_A, \mE_A, \mC_A)) \in \R^{h\times d^a}\,; \\
    &\mF_A^{S} = \text{SORT}_{D}(\text{SORT}_{F}(\mF_A^{G}))\,,
\end{align}
where $d^a$ is the atom feature dimension, $\text{SORT}_{F}$ and $\text{SORT}_{D}$ are fragment-based and node-degree-based sorts, respectively. Although Mamba~\cite{gu2023mamba} employs selective SSM (State Space Model) to overcome the time-invariant nature of the system like Transformer, 
we add \textbf{positional encoding} to the atom sequences fed into the Mamba system, to enhance the atomic position-aware capability of the MG module, as follows:
\begin{align}
\mF_A^{P} = \mF_A^{S}\|\mP_F\|\mP_D\,,
\end{align} 
where $\|$ denotes concatenation, $\mP_F$ encodes the fragment label for each atom, reflecting the fragment to which each atom belongs. $\mP_D$ encodes the intra-fragment positional information for atoms within the same fragment.

In order to better delve the structure information of graph, 
correlate and synchronize the graph and Mamba, we design a new \textbf{GSSM (GraphSSM) mechanism} as shown in Figure~\ref{fig:mamba-graph}, it allows the model to adaptively select relevant information from the graph context, with a Graph Information FeedForward workflow. Here, we feed the structural information of the graph (Adjacency matrix $\mA_G\in\R^{l\times l}$ and distance matrix $\mD_G\in\R^{l\times l}$) into the GSSM of Mamba. Specially, the GSSM can be achieved by remaking the SSM parameters ($\Amat$, $\Bmat$ and $\Delta$) as the new functions of the input data sequence, we express the process in Algorithm~\ref{alg:mamba_block}. The input of the Mamba block is $x=\mF_A^{P} \in \R^{h\times d^a}$, as $x$ passes through it, we perform the FragPool (MaxPool within each fragment) to obtain the fragment-level graph embedding:
\begin{align}
\mF_A^{M} = \text{FragPool}(\text{Mamba}(\mF_A^{P})) \in \R^{H\times d}.
\end{align}
In addition, two fragment-related structural losses are used for self-supervised constrained model training to facilitate structural inference in the MG module. Thanks to the optimized fragmentation procedure~\cite{nips_psma} that we use, we can enumerate all the trunks of the fragment-graph and use the output of Mamba block (full molecular graph information) to predict it and all subgraph's ids, as two multi-label prediction tasks, with the loss of $\mL_{structure}$ ($\mL_{s}$) and $\mL_{fragment}$ ($\mL_{f}$), respectively.

\begin{algorithm}[t!]
\caption{Mamba block with Graph SSM}
\label{alg:mamba_block}
\begin{algorithmic}[1]
\STATE \textbf{Input:} $x \in \R^{{b \times l\times d}}$ $\quad$\textbf{Output:} $y \in \R^{{b \times l\times d}}$ 
\STATE $\mA_G,\mD_G\in\R^{l \times l} \leftarrow \text{Adjacency and distance matrices}$ 
\STATE $x, z \in \mathbb{R}^{{b \times l\times d^h}} \leftarrow \text{Linear}(x), \text{Linear}(x)$
\STATE $x \in \mathbb{R}^{{b \times l\times d^h}} \leftarrow \text{SiLU}(\text{Conv1d}(x))$ 
\STATE $\Amat \in \mathbb{R}^{{d\times n}} \leftarrow \text{Parameter}$
\STATE $\Bmat, \Cmat \in \mathbb{R}^{{b \times l\times n}} \leftarrow \text{Linear}(x), \text{Linear}(x)$
\STATE $\Delta \in \mathbb{R}^{{b \times l\times d}} \leftarrow \text{Softplus}(\text{Linear}(x))$
\STATE $\Delta' \leftarrow (\Delta \odot \mA_G \odot \mD_G)\,$ 
\STATE $\overline{\Amat} \in \mathbb{R}^{{b \times l\times d\times n}} \leftarrow \exp (\sum_{i,j} \Delta'_{ij} \Amat_{ij})$ 
\STATE $\overline{\Bmat} \in \mathbb{R}^{{b \times l\times d\times n}} \leftarrow \sum_{i,j,k} \Delta'_{ij} B_{jk} x_{ki}$ 
\STATE $y \in \mathbb{R}^{{b \times l\times d^h}} \leftarrow \text{SSM}(\overline{\Amat},\overline{\Bmat}, \Cmat)(x)$ 
\STATE $y \in \mathbb{R}^{{b \times l\times d^h}} \leftarrow y \odot \text{SiLU}(z)$
\STATE \textbf{return} $y \in \mathbb{R}^{{b \times l\times d}} \leftarrow \text{Linear}(y) + x$
\end{algorithmic}
\end{algorithm}

\subsubsection*{Structure Collaborative Training.}
To train both fragment-level and atom-level graphs, we maximize the consistency between local structural information of molecular fragments and the global structural information. 
Specifically, we conduct the Softmax function on the $\mF_F$ from fragment-graph $\text{GNN}_F$ as a pseudo-label for the fragment distribution. Also, based on the output fragment embedding $\mF_A^{M}$ of MG module, we calculate the predicted labels. The objective function $\mL_{distribution}$ ($\mL_{d}$) is 
\begin{align}\label{eq:soft}
&\hat{\mF}_F=\text{SoftMax}(\tau \cdot \mF_F), \hat{\mF}_A =\text{SoftMax}(\tau \cdot \mF_A^{M}); \\
&\mL_{d} = -(\E_{\hat{f}_u^{F},u\in\mV_F} \text{log}(\hat{f}_u^A) + \E_{\hat{f}_u^{A},u\in\mV_A} \text{log}(\hat{f}_u^F)),
\end{align}
where $\tau$ is a temperature coefficient used to control the smoothness of the soft labels. $\mL_{d}$ is the cross-entropy loss designed to encourage
mutual supervision between the $\text{GNN}_F$ and MG modules.

\subsection{E-semantic Driven Structure \& Electron Fusing}
\subsubsection*{Molecular Electronic View Expression.} 
Molecular descriptors are employed to characterize the overall properties of molecules, and descriptors with prediction-relevant properties should be chosen, we statistics the descriptors that represent the molecules from the electronic view in Table~\ref{tab:elec_data}. Then the molecular electronic view expression is $\mD_E=\mD_S\|\mD_M\|\mD_Q\|\mD_C\in \R^{M\times 2}$. 
To integrate molecular structure and electronic information, we design an E-semantic masked training strategy, \ie, we randomly mask $\alpha$\% of the input sequence $\mD_E$ with the mask matrix $\mM$, then the MLPs serves as the E-encoder to convert the masked descriptor tokens $\mD_E^M$ into a sequence of tokens $\mS_E^M\in \R^{M\times d}$ for being further processed by Mamba-Transformer model.

\begin{table}[t!]
  \centering
  \begin{adjustbox}{width=0.85\linewidth}
  \begin{tabular}{l|ccc} 
    \hline
    {Type of Descriptors} &Notation &Dimension &Number \\  
    \hline
    E-state             &$\mD_S$ &2 &25 \\
    Molecular Property  &$\mD_M$ &2 &55 \\
    Quantum Chemical    &$\mD_Q$ &2 &7 \\
    Charge              &$\mD_C$ &2 &25 \\
    \hline
  \end{tabular}
  \end{adjustbox}
  \caption{The statistics of Electrochemical Descriptors.}
  \label{tab:elec_data}
\end{table}

\subsubsection*{Unified Mamba-Transformer Fuser.} For each molecule, we combine its structural and electronic representations.
We adopt the Mamba-Transformer (MT) backbone to integrate these two features, the MT module is shown in Figure~\ref{fig:model} (right). 
Here, the MT module is a state space augmented transformer variant to enhance modeling of long sequences (concatenation of structure and electronic embeddings). 
Based on the masked electronic feature $\mS_E^M$, we introduce a self-supervised mask prediction task, to enhance the interaction of structural and electronic features and to remove the effect of redundant electronic descriptors for downstream task prediction. The processes are 

\begin{align}
&\mU = \text{MT}(\mS_E^M \| \mF_A^M) \in \R^{(M+H)\times d}; \\
&\hat{\mS}_E^M = \text{MLP}(\mU[1:M,:]) \in \R^{M\times d}; \\
&\mL_{mask} = \frac{1}{\sum_{i=1}^{M} \mM_{i}} \sum_{i=1}^{M} \mM_{i} \left( (\mS_E^M)_{i} - (\hat{\mS}_E^M)_{i} \right)^2. \label{eq:mask}
\end{align}

\subsection{Optimization Objective} 
\subsubsection{Pretraining.} 
In our pretraining phase, we define four loss functions, each with specific optimization objectives. The overall objective is to minimize a weighted sum of these losses. The four losses are: structural distribution collaborative training loss $\mL_{d}$ of GNN$_F$ and MG module, two fragment-related structural losses $\mL_{s}$ and $\mL_{f}$ for MG module, masked E-semantic fusion training loss $\mL_{mask}$. The overall optimization objective combines these four loss functions with specific weights:
\begin{align}
    \mL_{\text{total}} = \lambda_d \mL_{d} + \lambda_s \mL_{s} + \lambda_f \mL_{f} + \lambda_{mask} \mL_{mask}
\end{align}
where $\lambda_d$, $\lambda_s$, $\lambda_f$, and $\lambda_{mask}$ are the weights for each loss component, respectively.

\subsubsection{Downstream Inference.} After pretraining, the learned encoders (GNN$_F$, MG module, E-encoder(MLPs) and MT fuser) are fine-tuned on specific downstream property prediction tasks. Note that the output of the first three encoders is concatenated and injected into MT fuser. Subsequently, a two-layer MLPs is attached to the MT module as a prediction head for downstream tasks. 
The fine-tuning process ensures that the model adapts to the specific requirements of each downstream task, leveraging the rich structural and electronic information captured during pretraining.

\section{Experiments}

In this section, we conduct comprehensive experiments to demonstrate the effcacy of our proposed method. The experiments are designed to analyze the method by addressing the following key questions:

\begin{itemize}[leftmargin=*]
    \item \textbf{Q1:} How does MOL-Mamba perform compared with state-of-the-art methods for molecular property prediction? 
    \item \textbf{Q2:} How do the fragment-graph GNN$_F$, the atom-level MG module and the MT fuser affect MOL-Mamba?
    \item \textbf{Q3:} Do $\mL_{distribution} (\mL_{d}), \mL_{structure} (\mL_{s}), \mL_{fragment} (\mL_{f})$ and $\mL_{mask}$ provide useful supervision for molecular structure and electronic learning during pretraining?
    \item \textbf{Q4:} How useful are graph node sorting strategy, position encoding and Mamba block with GSSM in MG Module?
    \item \textbf{Q5:} Quantitative mechanistic interpretation of the model.
\end{itemize}

\subsection{Experiment Settings}
\subsubsection*{Datasets.} 
We use the recently popular {GEOM}~\cite{data_geom} that contains 50k qualified molecules, for molecular pretraining, followed by~\cite{iclr_graphmvp, nips_evaluating}. 
For downstream tasks, we conduct experiments on 11 benchmark datasets from the MoleculeNet~\cite{data_moleculenet}, they involve physical chemistry, biophysics, physiology and quantum mechanics. Based on task type, 7  classification benchmarks (BBBP, Tox21, ClinTox, HIV, BACE, SIDER, MUV) and 4 regression benchmarks (FreeSolv, ESOL, Lipo, QM9) are included. Each dataset uses the recommended splitting method to divide data into training/validation/test sets with a ratio of 8:1:1. 

\begin{table*}[ht]
  \centering
  \begin{adjustbox}{width=0.9\linewidth}
  \begin{tabular}{lll|ccccccc} 
    \hline 
    &\multirow{2}{*}{Method} &\multirow{2}{*}{Venue} &BBBP &Tox21 &ClinTox &HIV &BACE &SIDER &MUV \\ 
    & & &2,039 &7,831 &1,478 &41,127 &1,513 &1,427 &93,087 \\ 
    \hline

{\multirow{4}{*}{\rotatebox{90}{supervised}}}&SchNet &{NIPS'17} &\textbf{84.8 (±2.2)} &77.2 (±2.3) &71.5 (±3.7) &70.2 (±3.4) &76.6 (±1.1) &53.9 (±3.7) &71.3 (±3.0) \\ 

&GIN &{ICLR'19} &65.8 (±4.5) &74.0 (±0.8)  &58.0 (±4.4) &75.3 (±1.9) &70.1 (±5.4) &57.3 (±1.6) &71.8 (±2.5)\\

&AttentiveFP &{JMC'19} &64.3 (±1.8) &\textbf{76.1 (±0.5)} &84.7 (±0.3) &\textbf{75.7 (±1.4)} &78.4 (±2.2) &60.6 (±3.2) &\textbf{76.6 (±1.5)} \\

&DMPNN &{NIPS'21} &71.2 (±3.8) &68.9 (±1.3) &\textbf{90.5 (±5.3)} &75.0 (±2.1) &\textbf{85.3 (±5.3)} &\textbf{63.2 (±2.3)} &76.2 (±2.8)\\ 

\hline

{\multirow{8}{*}{\rotatebox{90}{pretraining}}} &PretrainGNN &ICLR'20 &68.7 (±1.3) &78.1 (±0.6)  &72.6 (±1.5) &79.9 (±0.7) &84.5 (±0.7)   &62.7 (±0.8)  &81.3 (±2.1) \\

&GROVER &{NIPS'20} &69.5 (±0.1) &73.5 (±0.1) &76.2 (±3.7) &68.2 (±1.1)  &81.0 (±1.4)  &65.4 (±0.1)  &67.3 (±1.8) \\

&GEM &{NMI'22} &72.4 (±0.4) &78.1 (±0.1) &90.1 (±1.3) &80.6 (±0.9)  &85.6 (±1.1)  &67.2 (±0.4)  &81.7 (±0.5) \\

&GraphMVP &ICLR'22 &72.4 (±1.6) &74.4 (±0.2) &77.5 (±4.2) &77.0 (±1.2) &81.2 (±0.9) &63.9 (±1.2) &75.0 (±1.0)  \\

&MolCLR &NMI'22 &\underline{\textbf{73.6 (±0.5)}} &\underline{\textbf{79.8 (±0.7)}} &\textbf{93.2 (±1.7)}  &80.6 (±1.1)  &\textbf{89.0 (±0.3)} &\underline{\textbf{68.0 (±1.1)}} &\underline{\textbf{88.6 (±2.2)}}\\

&Uni-Mol &ICLR'23 &72.9 (±0.6) &79.6 (±0.5) &91.9 (±1.8) &\underline{\textbf{80.8 (±0.3)}} &85.7 (±0.2) &65.9 (±1.3)  &82.1 (±1.3)\\

&MOLEBLEND &ICLR'24 &73.0 (±0.8) &77.8 (±0.8) &87.6 (±0.7) &79.0 (±0.8) &83.7 (±1.4) &64.9 (±0.3)  &77.2 (±2.3)  \\

\rowcolor{gray!15}
&\multicolumn{2}{l|}{\bf MOL-Mamba (Ours)} &\textbf{75.0 (±0.2)} &\textbf{81.3 (±0.4)} &\underline{\textbf{92.7 (±1.1)}}  &\textbf{81.6 (±0.5)} &\underline{\textbf{86.4 (±0.3)}} &\textbf{68.3 (±0.9)} &\textbf{89.0 (±1.2)}\\
    \hline
  \end{tabular}
  \end{adjustbox}
  \caption{Results of state-of-arts on seven classification benchmarks. Mean and standard deviation of test ROC-AUC$\uparrow$  (\%) on each benchmark are reported. Best performing supervised and self-supervised/pretraining methods for each benchmark are marked as \textbf{bold}. Second best self-supervised methods are marked as \underline{\textbf{bold}}.
  }
  \label{tab:class}
\end{table*}

\begin{table}[t]
  \centering
  \begin{adjustbox}{width=1\linewidth}
  \begin{tabular}{l|cccc} 
    \hline 
    \multirow{2}{*}{Method} &FreeSolv &ESOL &Lipo &QM9 \\ 
    &642 &1,128 &4,200 &130,829 \\ 
    \hline

SchNet &3.22 (±0.76) &1.05 (±0.06) &0.91 (±0.10) &0.081 \\ 

GIN &2.76 (±0.18) &1.45 (±0.02) &0.85 (±0.07) &0.009 \\ 

AttentiveFP &\textbf{2.07 (±0.18)} &\textbf{0.88 (±0.03)} &0.72 (±0.00) &\textbf{0.008}  \\

DMPNN &2.18 (±0.91) &0.98 (±0.26) &\textbf{0.65 (±0.05)} &0.008 \\

\hline

PretrainGNN &2.76 (±0.00) &1.10 (±0.01)  &0.74 (±0.00) &0.009 \\

GROVER &2.27 (±0.05) &0.90 (±0.02)  &0.82 (±0.01) &0.010 \\

GEM &1.88 (±0.09) &0.80 (±0.03)  &0.66 (±0.01) &\underline{\textbf{0.007}} \\

GraphMVP &-  &1.03 (±0.03) &0.68 (±0.01) &-\\

MolCLR &2.20 (±0.20) &1.11 (±0.01) &0.65 (±0.08) &- \\

Uni-Mol &\underline{\textbf{1.62 (±0.04)}} &\underline{\textbf{0.79 (±0.03)}}  &\underline{\textbf{0.60 (±0.01)}}  &{\textbf{0.005}} \\

\rowcolor{gray!15}
\textbf{MOL-Mamba} &{\textbf{1.02 (±0.02)}} &{\textbf{0.63 (±0.01)}} &\textbf{0.53 (±0.01)} &\underline{\textbf{0.007}}\\
    \hline
  \end{tabular}
  \end{adjustbox}
  \caption{Results of state-of-arts on four regression benchmarks. Mean and standard deviation of test RMSE$\downarrow$  (for FreeSolv, ESOL, Lipo) or MAE$\downarrow$  (for QM9) are reported. Since the standard deviation of all results on QM9 is close to 0, we do not show them.
  }
  \label{tab:reg}
\end{table}

\begin{table}[t!]
  \centering
  \begin{adjustbox}{width=1\linewidth}
  \begin{tabular}{l|llll}
    \hline
    Model  &\#Params &FLOPs &Runtime(s/molecule) &ROC-AUC \\
    \hline
GIN &1.63M  &1.2M  &0.001 &74.0 (±0.8) \\
SchNet &1.86M &1.5M &0.0011 &77.2 (±2.3)  \\
Uni-Mol &47.1M &10M &0.0083 &79.6 (±0.5)  \\
MOLEBLEND &47.1M &11M &0.0091 &77.8 (±0.8)  \\

\rowcolor{gray!15}
\textbf{MOL-Mamba} &6.98M &4M &0.0012 &81.3 (±0.4) \\

    \hline
  \end{tabular}
  \end{adjustbox}
  \caption{Model complexity comparison on Tox21 dataset. 
  }
  \label{tab:eff}
\end{table}

\subsubsection*{Baselines.} We compare the proposed MOL-Mamba against multiple baselines, including both supervised and self-supervised/pretraining methods. 
\textbf{(1) Supervised GNN methods} include SchNet~\cite{nips_schnet}, GIN~\cite{iclr_gin}, AttentiveFP~\cite{jmc_attentionfp}, 
and DMPNN~\cite{nips_dmpnn}. SchNet models molecular quantum interactions using continuous-filter convolutional neural networks. GIN enhances the ability to learn complex graph structures through graph isomorphism and learnable aggregation functions. DMPNN and AttentiveFP are two variants of message passing neural networks~\cite{icml_mpnn}.
\textbf{(2) Pretraining methods} include PretrainGNN~\cite{iclr_pretraingnn}, GROVER~\cite{nips_grover}, GEM~\cite{nmi_gem}, GraphMVP~\cite{iclr_graphmvp}, MolCLR~\cite{nmi_molclr}, Uni-Mol~\cite{iclr_unimol}, and MOLEBLEND~\cite{iclr_moleblend}. 
\emph{PretrainGNN} introduces node-level self-supervised methods, such as context prediction and attribute masking, for pretraining GNNs.
\emph{GROVER} proposes a self-supervised graph transformer method for learning molecules through graph-level motif prediction.
\emph{GEM} employs a geometry-enhanced approach to capture molecular 3D spatial knowledge, including bond lengths, bond angles, and atomic distances.
\emph{GraphMVP} improves molecular structural representation via contrastive pretraining of both 2D topology and 3D molecular structures.
\emph{MolCLR} aligns substructures of similarly structured molecules to enhance molecular representation.
\emph{Uni-Mol} and \emph{MOLEBLEND} integrate 1D sequence, 2D topology, and 3D information to improve model performance and adaptability for various molecular tasks.

\subsubsection*{Implementation Details.} We follow the traditional metrics ROC-AUC$\uparrow$ and RMSE/MAE$\downarrow$~\cite{aaai_geomgcl} for the molecular property prediction classification and regression tasks respectively. And the performance metrics reported here are the averaged results of 10-fold cross-validation and the standard deviations. 
\emph{In the implementation}, molecular descriptors are calculated by the ChemDes package~\cite{chemdes}. We adopt 6-layer GIN~\cite{iclr_gin} as the fragment-graph GNN$_F$, and set a 6-layer SchNet~\cite{nips_schnet} as the atom-graph GNN$_A$ of the MG module, without any initialized parameters, and set 2-layer Mamba blocks. 
\emph{For pretraining}, we set the temperature coefficient in Eq.~\ref{eq:soft} $\tau=0.5$ , and we set the mask ratio as $\alpha=10$ (\%) for mask matrix $\mM$ in Eq.~\ref{eq:mask}. Based on the order of magnitude of each loss, we set different loss weights as follows, $\mL_d=\mL_s=\mL_{mask}=0.1, \mL_f=20.0$, respectively.
\emph{For pretraining and fine-tuning}, we employ the AdamW optimizer, the learning rate is set to 0.0001, and the batch size is 64, and the training is conducted 100 epochs, with the early stopping on the validation set. We develop all codes on a single NVIDIA RTX A5000 GPU.

\subsection{Performance Evaluation}
\subsubsection*{Overall Comparision (Q1).} 
Table~\ref{tab:class} and~\ref{tab:reg} present the experimental results of MOL-Mamba compared to competitive baselines, with the best results highlighted in bold. Our MOL-Mamba framework demonstrates superior performance on 8 out of 11 classification and regression benchmarks. For instance, it achieves a 75.0\% ROC-AUC with a standard deviation of 0.2 on the BBBP dataset, and the lowest mean MAE of 0.63 on the ESOL dataset. This underscores MOL-Mamba's strength as a robust pretraining framework that is both easy to implement and highly adaptable across various molecular domain tasks. Moreover, Table~\ref{tab:eff} shows that MOL-Mamba achieves comparable running time compared to GNNs (GIN and SchNet), \eg 0.0012 s/molecule vs. 0.001 s/molecule, but has a significant low-cost advantage in parameters and running time compared to GTs (Uni-Mol and MOLEBLEND), striking a good balance between complexity and performance. 
Comparing supervised methods, while SchNet excels on BBBP, it significantly underperforms on other datasets (notably 17.7\% below us on MUV). When trained with supervision only (Table~\ref{tab:loss}, Row 1), MOL-Mamba achieves 78.6\% SOTA on Tox21. As shown in Tables~\ref{tab:class} and~\ref{tab:reg}, MOL-Mamba consistently outperforms leading supervised baselines across datasets, demonstrating its robust ability to capture and integrate complex molecular information.

\begin{figure}[t]
\centering
\begin{subfigure}[b]{0.49\linewidth}
\centering
\begin{overpic}[width=1\linewidth]{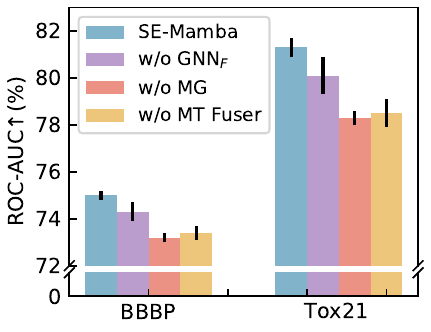} 
\end{overpic}
\end{subfigure}
\begin{subfigure}[b]{0.49\linewidth}
\centering
\begin{overpic}[width=1\linewidth]{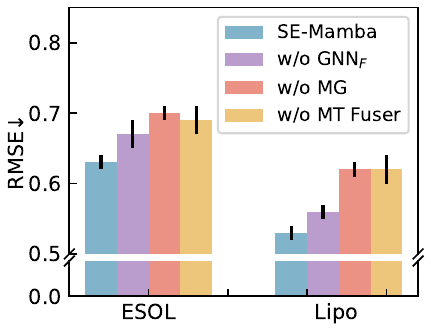} 
\end{overpic}
\end{subfigure} 
\caption{MOL-Mamba with different module settings on classification benchmarks BBBP and Tox21, regression benchmarks ESOL and Lipo, (lower is better for regression).
}
\label{fig:main_module}
\end{figure}

\begin{table}[t!]
  \centering
  \begin{adjustbox}{width=1\linewidth}
  \begin{tabular}{cccc|cc|cc}
    \hline
    $\mL_d$ &$\mL_s$ &$\mL_f$ &$\mL_{mask}$ &BBBP &Tox21 &ESOL &Lipo \\
    \hline
     - & - &- &-  & 72.5 (±0.4) &78.6 (±0.6) &0.70 (±0.03) &0.63 (±0.03)\\
     
     \dmark &-&-&- &73.4 (±0.3) &79.1 (±0.5) &0.68 (±0.02) &0.59 (±0.02)\\
     
     \dmark & \dmark &-&- &73.8 (±0.3) &80.5 (±0.5) &0.66 (±0.02) &0.58 (±0.02)\\
     
     \dmark & \dmark  &\dmark &- &74.2 (±0.3) & 80.9 (±0.4) &0.65 (±0.02) &0.56 (±0.02)\\

     \rowcolor{gray!15}
      \dmark & \dmark  &\dmark &\dmark &\textbf{75.0 (±0.2)} &\textbf{81.3 (±0.4)} &\textbf{0.63 (±0.01)} &\textbf{0.53 (±0.01)} \\

    \hline
  \end{tabular}
  \end{adjustbox}
  \caption{Mol-Mamba with different pretraining loss settings on classification benchmarks BBBP and Tox21, regression benchmarks ESOL and Lipo, (lower is better for regression).
  }
  \label{tab:loss}
\end{table}

\begin{table}[t!]
  \centering
  \begin{adjustbox}{width=0.95\linewidth}
  \begin{tabular}{l|cc|cc}
    \hline
    Model &BBBP &ToX21 &ESOL &Lipo \\
    \hline
\rowcolor{gray!15}
MG &\textbf{75.0 (±0.2)} &\textbf{81.3 (±0.4)} &\textbf{0.63 (±0.01)} &\textbf{0.53 (±0.01)}\\

w/o SORT & 73.2 (±0.4) & 79.5 (±0.5) & 0.66 (±0.03)&0.56 (±0.03)\\

w/o PE & 74.0 (±0.3) & 80.1 (±0.5) & 0.65 (±0.02) & 0.55 (±0.03)\\

w/o GSSM &73.0 (±0.3) &78.9 (±0.6) &0.67 (±0.03) &0.57 (±0.03) \\

    \hline
  \end{tabular}
  \end{adjustbox}
  \caption{MG module with different settings. ``SORT'' denotes the graph node sorting strategy, ``PE'' denotes positional encoding $\mP_F$ and $\mP_D$, ``GSSM'' refers to GraphSSM. 
  }
  \label{tab:mg}
\end{table}

\subsubsection*{Main Module Analysis (Q2).} 
Figure~\ref{fig:main_module} shows the critical role of MOL-Mamba's main modules. The absence of GNN\(_F\) module results in noticeable performance drops across datasets, underscoring its importance in capturing fragment-level information. Similarly, removing the Mamba-Graph (MG) module leads to accuracy decline, particularly in classification tasks like Tox21, emphasizing its role in atom-level structural reasoning. The MT Fuser module proves pivotal for integrating molecular structure and electronic data, as evidenced by decreased performance in both classification and regression tasks when omitted. These results demonstrate each component's unique contribution to MOL-Mamba's robust performance.

\subsubsection*{Self-supervised Loss Analysis (Q3).} Table~\ref{tab:loss} highlights the impact of pretraining loss settings on MOL-Mamba's performance. Without any self-supervised losses, \ie, the model is trained directly on the task dataset, the model underperforms, emphasizing the need for pretraining strategies. Introducing \(\mathcal{L}_d\) improves results by capturing key structural features. Adding \(\mathcal{L}_s\) further enhances performance, especially on Tox21, by boosting semantic-level reasoning. Incorporating \(\mathcal{L}_f\) provides further gains in BBBP and Lipo. 
The full model, integrating all loss components including \(\mathcal{L}_{\text{mask}}\), performs the best, effectively combining structural and electronic data for property predictions.

\begin{figure}[t]
\centering
\begin{overpic}[width=0.95\columnwidth]{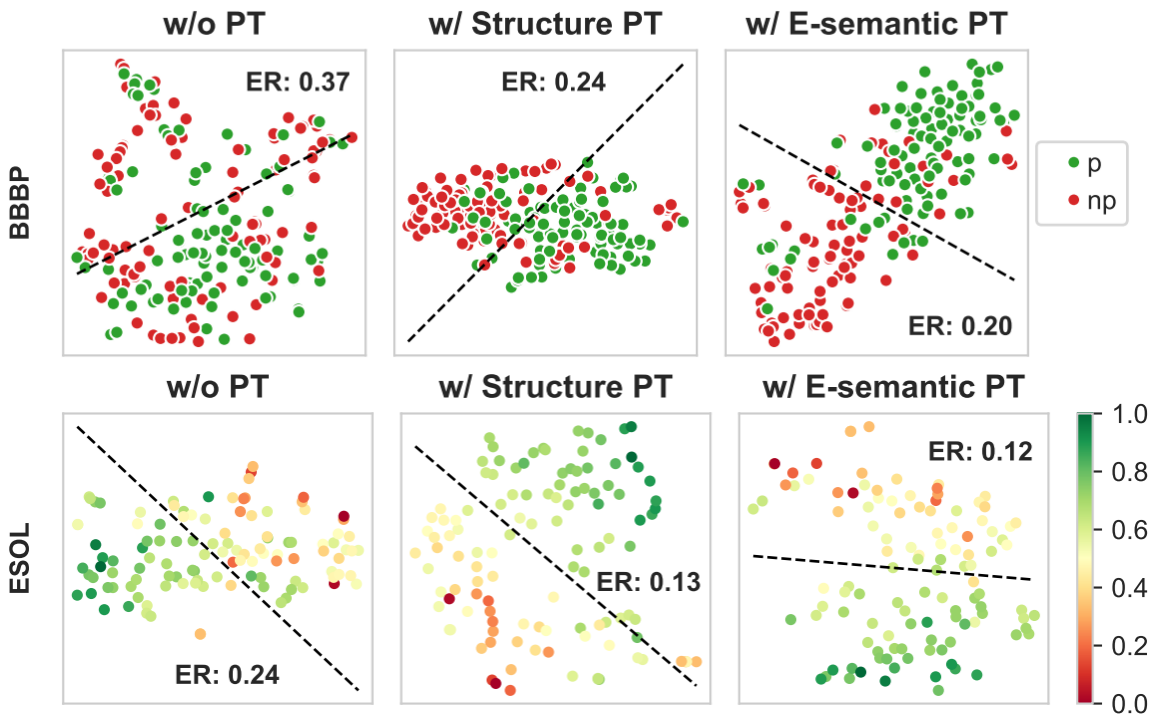} 
\end{overpic}
\caption{T-SNE visualization of feature separation of our pretraining (PT) methods on BBBP and ESOL. The ``Structure PT'' refers to pretraining the model with structural losses ($\mL_d, \mL_s, \mL_f$), the ``E-semantic PT'' pretrains the model with e-masked loss $\mL_{mask}$. ``ER'' is Error Rate.}
\label{fig:tsne}
\end{figure}

\subsubsection*{Mamba-Graph Module Analysis (Q4).} From Table~\ref{tab:mg}, the full MG configuration achieves optimal results across all benchmarks, underscoring its comprehensive design. Omitting the graph node sorting strategy ``SORT'') results in significant performance declines, particularly on the BBBP and Tox21 datasets, highlighting its essential role in node representation. Similarly, removing positional encoding (``PE'') reduces accuracy, indicating its importance for maintaining spatial information. The GraphSSM mechanism (``GSSM'') is crucial, as its absence causes the largest drop, especially in regression tasks like ESOL and Lipo. 
These findings demonstrate that each component is vital for the MG module's effectiveness in accurately predicting molecular properties.

\subsubsection*{Feature Separation Analysis (Q5).} Figure~\ref{fig:tsne} presents a t-SNE visualization illustrating the feature separation achieved by different pretraining (PT) methods on the BBBP and ESOL benchmarks. The error rate (ER), defined as the proportion of incorrectly classified points, is noted for each scenario. Without PT, the ER is highest at 0.37 for BBBP and 0.24 for ESOL, indicating poor separation between positive (p) and negative (np) samples. Introducing ``Structure PT'' significantly reduces the ER to 0.24 for BBBP and 0.13 for ESOL, demonstrating improved clustering due to enhanced structural understanding. The ``E-semantic PT'' further refines feature separation, achieving the lowest ERs of 0.20 and 0.12, respectively. This highlights the effectiveness of incorporating e-masked loss \(\mathcal{L}_{\text{mask}}\) in capturing complex semantic nuances, leading to more distinct and accurate classification boundaries.

\subsubsection*{Feature Visualization Analysis (Q5).} 
\begin{figure}[t]
\centering
\includegraphics[width=0.95\linewidth]{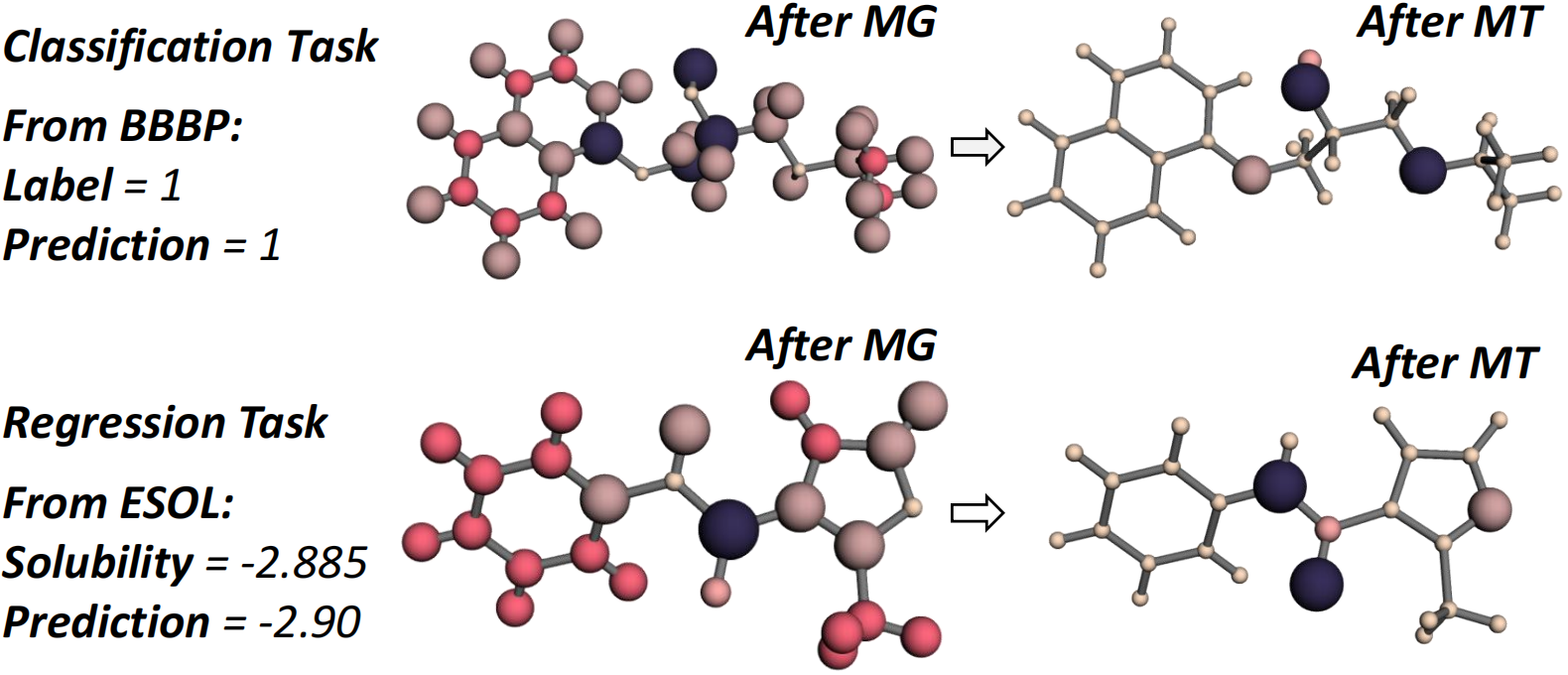}
\caption{Exemplary explanations for using MG and MT modules in molecular prediction tasks. The radius scales linearly with the node-feature sum. It highlights key structural and electronic adjustments.}
\label{fig:smiles}
\end{figure}

Visualization of feature weights provides insights into the relative importance of each input feature in the model’s decision-making process. 
In Figure~\ref{fig:smiles}, the feature transformations through the MG and MT modules are illustrated. For the classification task from BBBP, the MG module captures essential structural features, aligning predictions accurately with labels. The MT module further adjusts features by focusing on key electronic interactions, enhancing decision-making precision. In the regression task from ESOL, the MG module effectively models solubility characteristics, while the MT module refines these features, leading to highly accurate solubility predictions. This demonstrates the model’s ability to prioritize relevant molecular areas, such as reactive centers and stereochemistry. 
These transformations ensures that predictions are based on meaningful molecular patterns, enhancing the model's interpretability and reliability.

\section{Conclusion}
This paper introduces the MOL-Mamba framework, which significantly enhances molecular representation learning by integrating structural and electronic insights. Comprising hierarchical structural reasoning and a fusion encoder, MOL-Mamba employs innovative training strategies to achieve accurate molecular property predictions. Extensive experiments demonstrate its superior performance over state-of-the-art methods, offering a robust predictive tool for applications in drug discovery and materials science. 

\newpage
\section{Acknowledgments}
This work was supported by the National Natural Science Foundation of China (62272144,72188101, 62020106007, 62302139 and 22403087 for D.L.), the Major Project of Anhui Province (2408085J040, 202203a05020011),  the Fundamental Research Funds for the Central Universities (JZ2024HGTG0309, JZ2024AHST0337, and JZ2023YQTD0072), and the China Postdoctoral Science Foundation (2023TQ0343 for D.L.).

\bibliography{aaai25}

\end{document}